\documentclass[letterpaper, 10 pt, conference]{ieeeconf}  

\IEEEoverridecommandlockouts                              

\usepackage{graphics} 
\usepackage{epsfig} 
\usepackage{amsmath} 
\usepackage{amssymb}  
\usepackage{mathtools}
\usepackage{bbm}
\usepackage{xcolor, soul}
\sethlcolor{yellow}

\title{\LARGE \bf
Upper and Lower Bounds for End-to-End Risks \\ in Stochastic Robot Navigation}

\author{Apurva Patil$^1$ \and Takashi Tanaka$^2$ 
\thanks{*This work is supported by Lockheed Martin Corporation and FOA-AFRL-AFOSR-2019-0003.}
\thanks{$^{1}$ Walker Department of Mechanical Engineering, University of Texas at Austin. {\tt\small apurvapatil@utexas.edu}.
         $^{2}$Department of Aerospace Engineering and Engineering Mechanics, University of Texas at Austin.
        {\tt\small ttanaka@utexas.edu}. }
}
\begin{document}
\maketitle
\thispagestyle{empty}
\pagestyle{empty}
\begin{abstract}
We present novel upper and lower bounds to estimate the collision probability of motion plans for autonomous agents with discrete-time linear Gaussian dynamics. Motion plans generated by planning algorithms cannot be perfectly executed by autonomous agents in reality due to the inherent uncertainties in the real world. Estimating collision probability is crucial to characterize the safety of trajectories and plan risk optimal trajectories. Our approach is an application of standard results in probability theory including the inequalities of Hunter, Kounias, Fr\'echet, and Dawson. Using a ground robot navigation example, we numerically demonstrate that our method is considerably faster than the na\"ive Monte Carlo sampling method and the proposed bounds are significantly less conservative than Boole’s bound commonly used in the literature.
\end{abstract}
\section{Introduction}\label{Sec: Preliminary}
\subsection{Motivation}
Motion plans for mobile robots in obstacle-filled environments can be generated by autonomous trajectory planning algorithms \cite{lavalle2006planning}. For real-time implementations, robots are typically equipped with a trajectory tracking controller to mitigate the effects of modeling errors, disturbances, and measurement noises. Since the planned trajectory cannot be tracked perfectly in stochastic environments, collisions with obstacles occur with a nonzero probability in general, even though the planned trajectory is collision-free. To address this issue, risk-aware motion planning has received considerable attention over the years \cite{pepy2006safe}, \cite{blackmore2011chance}, \cite{strawser2018approximate}. Optimal planning under set-bounded uncertainty provides some solutions against worst-case disturbances \cite{majumdar2013robust}, \cite{lopez2019dynamic}. However, in many cases, modeling uncertainties with unbounded distributions, such as Gaussian distributions, has a number of advantages over a set-bounded approach \cite{blackmore2011chance}. In the case of unbounded uncertainties, in general, it is difficult to guarantee safety against all realizations of noise. This motivates for an efficient risk estimation technique that can both characterize the safety of trajectories and be embedded in the planning algorithms to allow explicit trade-offs between control optimality and safety. Assuming that a planned trajectory with a finite length in a known configuration space is given, we present several upper and lower bounds for the collision probability while tracking the trajectory. This probability will hereafter be called the \textit{end-to-end probability of failure}. The analysis in this paper assumes that the system dynamics are discrete-time. However, in Section \ref{subsection: Continuous-Time End-to-End Risks}, we also study the performance of our discrete-time risk bounds in the continuous-time setting as the underlying discretization is refined.\par
The paper is organized as follows: in Section \ref{Sec: Preliminary}, we formally define the problem of end-to-end risk analysis, review state-of-the-art literature, and state the contributions of this paper. In Section \ref{Sec: PROBABILITY BOUNDS}, we review probability inequalities from the literature based on which we derive upper and lower bounds of the end-to-end risks in Section \ref{Sec: END-TO-END RISK ANALYSIS}. In Section \ref{Sec: EXAMPLE}, we demonstrate the results of our analysis using a ground robot navigation example. Finally, Sections \ref{Sec: DISCUSSION} and \ref{Sec: CONCLUSION} are devoted to discussion and conclusion.\par
\subsection{Problem Formulation}\label{Sec: PROBLEM FORMULATION}
Let $\chi\subseteq \mathbb{R}^d$ be a known configuration space, where $d\in\mathbb{N}$, $d\geq2$. Let $\chi_{obs}\subset\chi$, $\chi_{free}=\chi\backslash\chi_{obs}$ and $\chi_{goal}\subset \chi_{free}$ be the obstacle region, obstacle-free region, and target region, respectively. Given an initial position $x_0^{plan}\in\chi_{free}$ of the robot, a planning algorithm generates a trajectory $\{x_t^{plan}\}_{t=0, 1, \hdots, T}$ by designing a finite, optimal sequence of control inputs $\{u_t^{plan}\}_{t=0, 1, \hdots, T-1}$ such that the end point of the trajectory satisfies $x_T^{plan}\in\chi_{goal}$. 
We call the finite sequence $\{x_t^{plan}\}_{t=0, 1, \hdots, T}$ the \textit {planned trajectory}, which satisfies    
\begin{equation*}\label{planned dynamics}
    x_{t+1}^{plan}=A_t x_t^{plan}+B_t u_t^{plan},\qquad t\in\{0,1,\hdots,T-1\}.
\end{equation*}
Let $x_t^{true}$ be the actual position of the robot during the execution of the plan and $u_t^{true}$ be the control input applied at time step $t$. In order to compensate for the effects of motion and sensing uncertainties, we assume the robot executes the planned trajectory in a closed-loop fashion \cite{stengel1994optimal}. We call the finite sequence $\{x_t^{true}\}_{t=0, 1, \hdots, T}$ the \textit {executed trajectory}. For the analysis purpose, in this paper, the system dynamics and the control policy are assumed to be linear. For nonlinear systems, we assume that the dynamics are linearized around the planned trajectories. Such an approach is shown to be effective in many control applications \cite{schouwenaars2001mixed}, \cite{schouwenaars2001plume}. We assume that the executed trajectory satisfies
\begin{equation*}\label{true dynamics}
\begin{split}
    x_{t+1}^{true}=A_t x_t^{true}+B_t u_t^{true}+w_t,\qquad w_t\sim\mathcal{N}\left(0,W_t\right),
\end{split}
\end{equation*}
at each time step $t\in\{0,1,\hdots,T-1\}$, where $w_t$ is a Gaussian white noise that models the motion uncertainty. The sensor model is given by
\begin{equation}\label{sensor model}
\begin{split}
    y_t=C_t x_t^{true}+v_t,\qquad v_t\sim\mathcal{N}\left(0,V_t\right),
\end{split}
\end{equation}
for $t\in\{0,1,\hdots,T-1\}$,
where $v_t$ is a Gaussian white noise that models the noise in the measurements. The sequence $\{\left(C_t, V_t\right)\}_{t=0,1,\hdots,T-1}$ is assumed to be known \textit{\textit{a priori}}.\par
Since our main focus is to evaluate the risk of a given trajectory plan, we assume that the trajectory is already provided. If $E_t$ represent the event that the robot collides with the obstacles at time step $t$ while tracking the planned trajectory, the end-to-end probability of failure in the trajectory tracking phase can be formulated as
\begin{equation}\label{end-to-end probability of failure} 
 P\left(\bigcup\limits_{t=0}^{T}x_t^{true}\in\chi_{obs}\right)=P\left(\bigcup\limits_{t=0}^{T} E_t\right).  
 \end{equation}\par
\subsection{Literature Review}
Monte Carlo and other sampling-based methods \cite{blackmore2010probabilistic}, \cite{janson2018monte}, \cite{calafiore2006scenario} provide accurate estimates of (\ref{end-to-end probability of failure}) by computing the ratio of the number of simulated executions that collide with obstacles. However, these methods are often expensive in computation due to the need for a large number of simulation runs to obtain reliable estimates and are cumbersome to embed in planning algorithms. \par
Various analytical approaches also have been proposed in the literature. In general, $\{E_t\}_{t=0, 1, \hdots, T}$ are statistically dependent events. Using the law of total probability, we can reformulate (\ref{end-to-end probability of failure}) as
\begin{equation}\label{probability of failure}
P\left(\bigcup\limits_{t=0}^{T} E_t\right)=1-\prod\limits_{t=0}^{T}P\left(E^c_t|E^c_{t-1}, \hdots, E^c_{0}\right)
\end{equation}
where $E_t^c$ represents the event that the robot is collision-free at time step $t$. It is challenging to compute (\ref{probability of failure}) exactly because it requires evaluating integrals of multivariate distributions over non-convex regions. One approach to estimate (\ref{probability of failure}) is to assume that the event $E_t$ is independent of other events or depends only on $E_{t-1}$ \cite{strawser2018approximate}, and subsequently approximate (\ref{probability of failure}) as $1- \prod\limits_{t=0}^{T}P\left(E^c_t\right)$ or $1-\prod\limits_{t=0}^{T}P\left(E^c_t|E^c_{t-1}\right)$,  respectively. However, these assumptions do not hold in general, and can result in overly conservative estimates or can even underestimate the failure probability.\par
Another popular approach is to use Boole's inequality (also called Bonferroni's first-order inequality) to compute an upper bound of (\ref{end-to-end probability of failure}) \cite{blackmore2011chance}, \cite{ono2008efficient}, \cite{ono2015chance}, \cite{blackmore2009convex}. The inequality is given as
 \begin{equation}\label{Bonferroni's first order upper bound}
     P\left(\bigcup\limits_{t=0}^{T} E_t\right)\leq \text{min}\left(\sum\limits_{t=0}^{T}P\left(E_t\right),1\right). 
 \end{equation}
This approach again ignores the dependency among events $\{E_t\}_{t=0, 1, \hdots, T}$ and can result in overly conservative estimates, especially as the time discretization is refined. When these estimates are used in planning risk-optimal paths, the algorithms might either find overly conservative paths or fail to find a feasible path, even if one exists.\par

In contrast to the previous approaches, an approach presented in \cite{patil2012estimating} accounts for the fact that the distribution of the state at each time step along the trajectory is conditioned on the previous time steps being collision-free. It truncates the estimated distributions of the robot's positions with respect to obstacles and approximates the truncated distributions as Gaussians. However, the Gaussianity is inexact and this approximation leads to an estimate that might not remain statistically consistent \cite{frey2020collision}.\par 

\subsection{Contributions}
The main contributions of the paper are summarized as follows:
In this work, we account for the fact that the events of collision at different time-steps $\{E_t\}_{t=0, 1, \hdots, T}$ are statistically dependent. Unlike \cite{patil2012estimating}, we compute the joint distribution of the entire robot trajectory without attempting to approximate the conditional state distributions. Using this joint trajectory distribution, we derive both upper and lower bounds for the end-to-end probabilities of failure. Our upper bounds are considerably tighter than the estimates obtained by Boole's inequality commonly used in the literature. The lower bounds, on the other hand, are useful for predicting how conservative the computed upper bounds are. Further, we show, in simulation, the validity and performance of our bounds, using a ground robot navigation example. We demonstrate that our method is considerably faster than the Monte Carlo sampling method. The approach presented in this paper is quite general and can be applied to estimate the discrete-time risks in stochastic navigation of any motion plan generated by an arbitrary planning algorithm.

 \section{PROBABILITY BOUNDS}\label{Sec: PROBABILITY BOUNDS}
In this section, we summarize first and second-order inequalities for the probability of union of events. These inequalities require computation of the terms $P\left(E_t\right)$ and $P\left(E_s\bigcap E_t\right)$, $s\neq t$, which are often easy to calculate. We define
\begin{equation*}
E\coloneqq \bigcup\limits_{t=0}^{T} E_t,\quad p_t\coloneqq P\left(E_t\right), \quad p_{s, t}\coloneqq P\left(E_s\bigcap E_t\right)  
\end{equation*}
and
\begin{equation}\label{S_1,S_2}
S_1\coloneqq \sum\limits_{0\leq t\leq T} p_t, \quad S_2\coloneqq \sum\limits_{0\leq s<t\leq T} \!\!\!p_{s, t}.
\end{equation}
Following are the upper bounds for the probability of union of events:
\begin{itemize}
\item Kwerel's upper bound:
\begin{equation}\label{Kwerel's upper bound}
P\left(E\right)\leq \text{min}\left(S_1-\frac{2}{T+1}S_2,1\right).   
\end{equation}
Here, $T+1$ is the total number of events in the union. This inequality was proved by Kwerel \cite{kwerel1975most}, \cite{kwerel1975bounds} as well as Sathe, Pradhan and Shah \cite{sathe1980inequalities}. It is the closest upper bound for the probability of the union of events based on the knowledge of $S_1$ and $S_2$. 
\item Kounias' upper bound:
\begin{equation} \label{Kounias' upper bound}
P\left(E\right)\leq \text{min}\left(S_1-\underset{0\leq s\leq T} {\text{max}} \sum\limits_{0\leq t\leq T,\,t\neq s} \!\!\!\!\!p_{s, t},1\right).
\end{equation}
\item Hunter's upper bound:
\begin{equation} \label{Hunter's upper bound (tight)}
P\left(E\right)\leq \text{min}\left(S_1-\underset{\tau}{\text{max}}\sum\limits_{\left(s,t\right):e_{s, t}\in \tau} \!\!\!\!\!p_{s, t},1\right).
\end{equation}
Here, $\tau$ is a spanning tree of the graph whose vertices are $E_0, E_1, \hdots, E_T$, with $E_s$ and $E_t$ joined by an edge $e_{s, t}$ if and only if $E_s\bigcap E_t\neq\emptyset$. Kruskal's minimum spanning tree algorithm \cite{kruskal1956shortest} can be used to find the $\tau$ which attains the maximum of $\sum p_{s, t}$. Kounias' inequality (\ref{Kounias' upper bound}) uses the maximum of $\sum p_{s, t}$ over only a subset of all spanning trees. Hence, Hunter's bound is sharper than Kounias' bound. Also, Hunter’s bound is always at least as good as Kwerel's upper bound (\ref{Kwerel's upper bound}).\par
In this work, we also compute a suboptimal Hunter's bound choosing a particular spanning $\tau$ having edges $e_{0, 1}, e_{1, 2}, \hdots, e_{T-1, T}$: 
\begin{equation} \label{Hunter's upper bound}
P\left(E\right)\leq \text{min}\left(S_1-\sum\limits_{1\leq t\leq T}p_{t-1, t},1\right).
\end{equation}
Compared to (\ref{Hunter's upper bound (tight)}), the bound in (\ref{Hunter's upper bound}) is cheaper in computation. It also possesses the time-additive structure similar to Boole's bound (\ref{Bonferroni's first order upper bound}); hence, this bound could be embedded in the risk-aware motion planning framework.
\end{itemize}
Following are the lower bounds for the probability of union of events:
\begin{itemize}
\item Fr\'echet's lower bound: $P\left(E\right)\geq\underset{0\leq t\leq T}{\text{max}}\,p_t.$
\item Bonferroni's second-order lower bound \cite{prekopa1988boole}:
\begin{equation} \label{Bonferroni's second order lower bound}
 P\left(E\right)\geq \text{max}\left(0,S_1-S_2\right).
\end{equation}
\item Dawson and Sankoff's lower bound \cite{dawson1967inequality}: If $S_1>0$,  
\begin{equation*}\label{Dawson and Sankoff's lower bound}
P\left(E\right)\geq \text{max}\left(0,\frac{2}{k+1}S_1-\frac{2}{k\left(k+1\right)}S_2\right),   
\end{equation*}
where $k-1$ is the integer part of $2S_2/S_1$. It is the closest lower bound for the probability of union of events based on the knowledge of $S_1$ and $S_2$. This optimality was proved by Galambos \cite{galambos1977bonferroni}. 
\end{itemize}
\section{END-TO-END RISK ANALYSIS} \label{Sec: END-TO-END RISK ANALYSIS}
In this section, we present a method to estimate the end-to-end risks based on the bounds given in Section \ref{Sec: PROBABILITY BOUNDS}.  
\subsection{Trajectory Tracking Controller}\label{subsection:Trajectory Tracking Controller} 
During an actual execution of the planned trajectory, the robot will likely deviate from the plan due to motion and sensing uncertainties. In order to compensate for these uncertainties, we assume the robot executes the plan using a linear feedback controller. In this paper, we use the LQG controller \cite{stengel1994optimal} and present here the derivation of the control policy briefly. For $t\in\{0,1,\hdots,T\}$, let
\begin{equation*}\label{definition xt}
    x_t\coloneqq x_t^{true}-x_t^{plan} \text{ and }
    u_t\coloneqq u_t^{true}-u_t^{plan}
\end{equation*}
be the deviation of the robot from the planned trajectory. The deviation is governed by
\begin{equation} \label{deviation from the planned path}
\begin{aligned}
    x_{t+1}& =A_t x_t+B_t u_t+w_t,\qquad w_t\sim\mathcal{N}\left(0,W_t\right),\\
    y_t& =C_t x_t+v_t,\qquad v_t\sim\mathcal{N}\left(0,V_t\right).
  \end{aligned}
\end{equation}
We assume the robot is initially at $x_0^{plan}$ and thus $x_0=0$. 
Let $\phi_t: \{y_t\}_{t=0, 1, \hdots, t}\rightarrow u_t$ be a feedback policy at time $t$, based on observations $\{y_t\}_{t=0, 1, \hdots, t}$. The optimal control law can be derived solving the following optimization problem: 
\begin{equation}\label{trajectory tracking problem}
\underset{\{\phi_t\}_{t=0, 1, \hdots, T-1}}{\arg\min}\quad\sum_{t=0}^{T-1}\mathbb{E}\left[\|x_{t+1}\|^2_{Q_t}+\|u_t\|^2_{R_t}\right],
\end{equation} 
where $Q_t\succeq0$, $R_t\succ0$ and $t\in\{0,1,\hdots,T-1\}$. 
The solution of Problem (\ref{trajectory tracking problem}) can be obtained using the separation principle. The optimal controller is given as
\begin{equation}\label{optimal controller}
    u_t=F_t\hat{x}_{t|t},\qquad t\in\{0,1,\hdots,T-1\},
\end{equation}
where $F_t$ are the LQR gains and $\hat{x}_{t|t}$ are the state estimates based on the measurements $\{y_t\}_{t=0, 1, \hdots, t}$. The LQR gains are computed as
\begin{equation*}\label{LQR gain}
    F_t=-\left(B_t^TH_t B_t+R_t\right)^{-1}B_t^TH_t A_t,
\end{equation*}
where $H_t$ is obtained using the backward Riccati recursion:
\begin{equation*}
\begin{split}
    H_{T\!-\!1} & =\!Q_{T\!-\!1},\\
    H_{t\!-\!1} & =\!Q_{t\!-\!1}\!+\!A_t^T\!H_tA_t\!-\!A_t^T\!H_tB_t\!\left(\!B_t^T\!H_tB_t\!+\!R_t\!\right)\!^{\!-\!1}\!B_t^T\!H_tA_t.
\end{split}
\end{equation*}
The state estimates $\hat{x}_{t|t}$ are determined by the Kalman filter. Let $P_{t|t-1}$ and $P_{t|t}$ be the \textit{a priori} and \textit{a posteriori} covariances, respectively, at time $t$. 
The \textit{a priori} and \textit{a posteriori} state estimates are computed as
\begin{equation}\label{Kalman estimates}
\begin{split}
    \hat{x}_{t|t-1}& =A_{t-1}\hat{x}_{t-1|t-1}+B_{t-1}u_{t-1},\quad\hat{x}_{0|0}=0\\
    \hat{x}_{t|t}& =\hat{x}_{t|t-1}+G_t\left(y_t-C_t\hat{x}_{t|t-1}\right),
\end{split}
\end{equation}
where $G_t$ are the Kalman gains that are evaluated as 
\begin{equation*}\label{Kalman gain}
    G_t=P_{t|t-1}C_t^T\left(C_t P_{t|t-1}C_t^T+V_t\right)^{-1}.
\end{equation*}
$P_{t|t-1}$ and $P_{t|t}$ are computed using the forward Riccati recursion with the initial condition $ P_{0|0}=0$:
\begin{equation*}
\begin{split}
      P_{t|t-1}& =A_{t-1}P_{t-1|t-1}A^T_{t-1}+W_{t-1},\\
      P_{t|t}^{-1}& =P_{t|t-1}^{-1}+C_t^TV_t^{-1}C_t.
\end{split}
\end{equation*}
\subsection{Distribution of the Closed-Loop Trajectory}\label{subsection:Distribution of the Closed-Loop Trajectory}
Combining (\ref{deviation from the planned path}), (\ref{optimal controller}) and (\ref{Kalman estimates}), the state deviation $x_t$ and its \textit{a priori} estimate $\hat{x}_{t|t-1}$ jointly evolve as \cite{patil2012estimating}:
\begin{equation*}\label{new closed-loop sys}
\begin{split}
\overline{x}_{t+1}=\overline{A}_t\overline{x}_t
+\overline{w}_t,&\qquad\overline{w}_t\sim\mathcal{N}\left(0,\overline{W}_t\right)
\end{split}
\end{equation*}
where
\begin{equation*}\label{xbar_t,wbar_t}
    \overline{x}_t\coloneqq\begin{bmatrix}
    x_t\\
    \hat{x}_{t|t-1}
    \end{bmatrix},\quad
    \overline{w}_t\coloneqq\begin{bmatrix}
    B_t F_t G_t v_t+w_t\\
    \left(A_t+B_t F_t\right)G_t v_t
    \end{bmatrix},
\end{equation*}

\begin{equation*}\label{A_t}
    \overline{A}_t\coloneqq\begin{bmatrix}
    A_t+B_t F_t G_t C_t&B_t F_t\left(I-G_t C_t\right)\\
    \left(A_t+B_t F_t\right)G_t C_t&\left(A_t+B_t F_t\right)\left(I-G_t C_t\right)
    \end{bmatrix}
    \end{equation*}
and
\begin{equation*}\label{W_t}
    \begin{split}
    \overline{W}_t\coloneqq&\\
    &\!\!\!\!\!\!\!\!\!\!\!\!\!\!\!\!\!\!\!\begin{bmatrix}
    \!B_t F_t G_t V_t G_t^T\!F_t^T\!B_t^T\!\!+\!W_t &\!\! B_t F_t G_t V_t G_t^T\!\left(\!A_t\!+\!\!B_t F_t\!\right)^{\!T}\\
    \!\left(\!A_t\!+\!\!B_t F_t\!\right)\!G_t V_t^T\!G_t^T\!F_t^T\!B_t^T & \!\! \left(\!A_t\!+\!\!B_t F_t\!\right)\!G_t V_t G_t^T\!\!\left(\!A_t\!+\!\!B_t F_t\!\right)^{\!T}\!
    \end{bmatrix}.
\end{split}
\end{equation*}
Stacking $\overline{x}_{t}$ for all time steps, we can write the equation of the \textit{closed-loop trajectory} as
\begin{equation*}\label{closed-loop trajectory}
    \overline{x}_{traj}=M\overline{x}_0+N\overline{w}_{traj},\qquad\overline{w}_{traj}\sim\mathcal{N}\left(0,\underset{0 \leq t \leq T-1}{\text{diag}}\overline{W}_t\right),
\end{equation*}
where
\begin{equation*}\label{xbar_traj,wbar_traj}
\overline{x}_{traj}\coloneqq\begin{bmatrix}
\overline{x}_0\\
\overline{x}_1\\
\overline{x}_2\\
\vdots\\
\overline{x}_T
\end{bmatrix}, 
\overline{w}_{traj}\coloneqq\begin{bmatrix}
\overline{w}_0\\
\overline{w}_1\\
\overline{w}_2\\
\vdots\\
\overline{w}_{T-1}
\end{bmatrix},
M\coloneqq\begin{bmatrix}
I\\
\overline{A}_0\\
\overline{A}_1\overline{A}_0\\
\vdots\\
\overline{A}_{T-1}\hdots\overline{A}_0\\
\end{bmatrix},
\end{equation*}
and
\begin{equation*}\label{N}
N\coloneqq\begin{bmatrix}
0&0&\hdots&0\\
I&0&\hdots&0\\
\overline{A}_1&I&\hdots&0\\
\vdots&\vdots&\ddots&\vdots\\
\overline{A}_{T-1}\hdots\overline{A}_1&\overline{A}_{T-1}\hdots\overline{A}_2&\hdots&I\end{bmatrix}.
\end{equation*}
Assuming $\overline{x}_0\sim\mathcal{N}\left(0,0\right)$, the distribution of the closed-loop trajectory can be written as
$
    \overline{x}_{traj}\sim\mathcal{N}\left(0,\overline{X}_{traj}\right)$
where
\begin{equation}\label{X_traj}
   \overline{X}_{traj}\coloneqq N\underset{0 \leq t \leq T-1}{\text{diag}}\left(\overline{W}_t\right)N^T. 
\end{equation}
\subsection{Computation of the Bounds}
\label{sec:computation of bounds}
We can obtain the exact end-to-end probability of failure by integrating the distribution of the closed-loop trajectory  over finite regions. However, as stated earlier, obtaining an integral in a high dimensional space is a computationally expensive problem. Instead, we make use of the probability inequalities listed in Section \ref{Sec: PROBABILITY BOUNDS} to obtain bounds for the end-to-end probability of failure. The main task in evaluating these bounds is to compute the univariate probabilities $p_t$, $t\in\{0,1,\hdots,T\}$ and bivariate joint probabilities $p_{s, t}$, $s,t\in\{0,1,\hdots,T\}$, $s\neq t$. These are computed from the distribution of $x_t^{true}$ and $\begin{bmatrix}x_s^{true}&x_t^{true}\end{bmatrix}^T$ respectively:
\begin{equation*}\label{individual distribution}
x_t^{true}\sim\mathcal{N}\left(x_t^{plan},X_t\right),
\end{equation*}
\begin{equation*}\label{joint distribution}
\begin{bmatrix}x_s^{true}&x_t^{true}\end{bmatrix}^T\sim\mathcal{N}\left(\begin{bmatrix}x_s^{plan}&x_t^{plan}\end{bmatrix}^T,X_{st}\right), 
\end{equation*}
where $X_t$ and $X_{st}$ are obtained by marginalizing $\overline{X}_{traj}$. Then,
 \begin{equation}\label{p_t}
     p_t=\int_{\chi_{obs}}\mathcal{N}\left(x_t^{plan},X_t\right)dx_t^{true}
 \end{equation}
 and
 \begin{equation}\label{p_st}
     p_{s, t}=\int_{\chi_{obs}}\!\int_{\chi_{obs}}\!\!\!\!\!\mathcal{N}\left(\begin{bmatrix}x_s^{plan}&x_t^{plan}\end{bmatrix}^T,X_{st}\right)dx_s^{true}dx_t^{true}.
 \end{equation}
If the obstacles in the configuration space are polyhedral, the method used in this work for the computation of the integrals (\ref{p_t}) and (\ref{p_st}) is summarized in the appendix. 
\section{SIMULATION RESULTS} \label{Sec: EXAMPLE}
In this Section, we demonstrate, in simulation, the validity of our bounds for the end-to-end risks, using a ground robot navigation example. The configuration space is $\chi=[0, 1]\times[0, 1]$ and the planned trajectory is assumed to satisfy
\begin{equation*}\label{eg planned dynamics}
    x_{t+1}^{plan}=x_t^{plan}+u_t, \qquad t\in\{0,1,\hdots,T-1\}. 
\end{equation*}
The executed trajectory $\{x_t^{true}\}_{t=0, 1, \hdots, T}$ satisfies the linearized robot dynamics, 
\begin{equation}\label{eg true dynamics}
\begin{split}
    x_{t+1}^{true}=x_t^{true}+u_t^{true}+w_t, &\qquad w_t\sim\mathcal{N}\left(0,W_t\right)
\end{split}
\end{equation}
for $t\in\{0,1,\hdots,T-1\}$, where  $W_t=\|x_{t+1}^{plan}-x_t^{plan}\|Z$, with $Z=10^{-3}\times I$ ($I$ is a $2\times 2$ identity matrix). (\ref{eg true dynamics}) is a natural model for ground robots whose location uncertainty grows linearly with the distance traveled. The sensor model is given as per (\ref{sensor model}). $u_t^{true}$ is computed using the LQG feedback control policy to minimize the deviation of the robot from the planned trajectory as explained in Section \ref{subsection:Trajectory Tracking Controller}. \par
\begin{figure}[tbhp]
    \centering
      \begin{tabular}{c c}
      \includegraphics[scale=0.3]{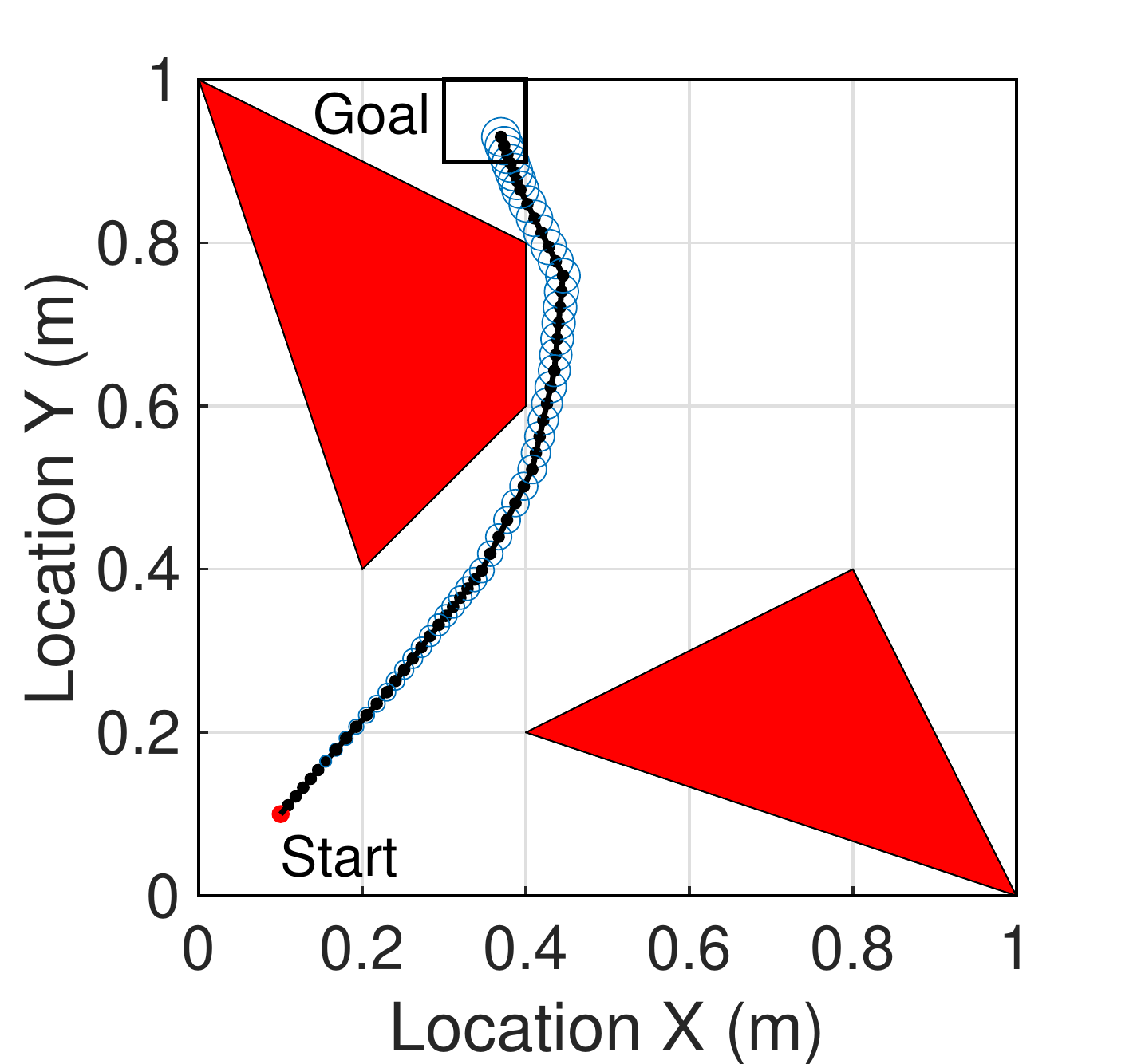} &\!\!\!\!\!\!\!\!\!\!\!\!\!\!\!\includegraphics[scale=0.3]{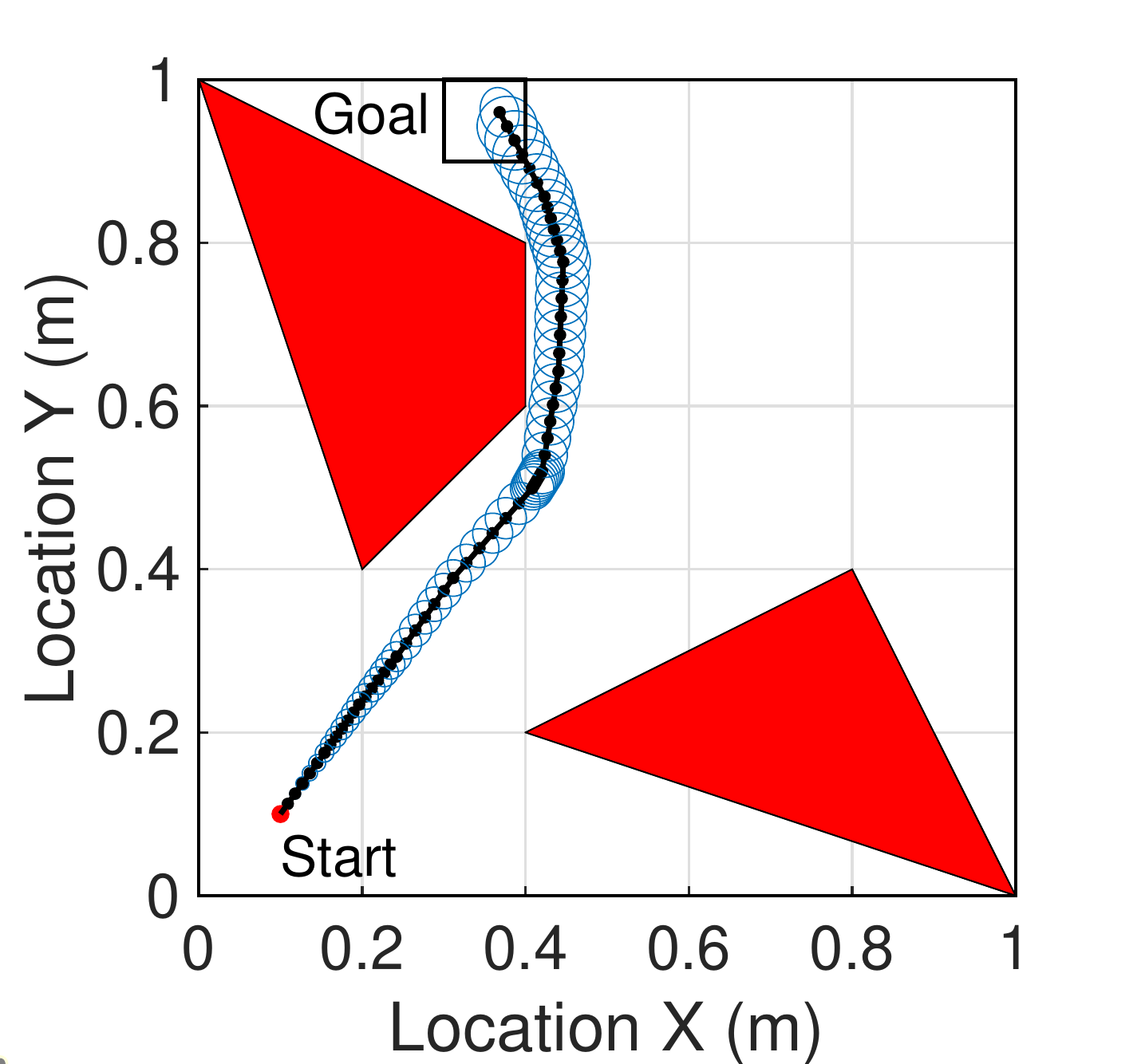} \\
      (a)&(b)\\
        \includegraphics[scale=0.3]{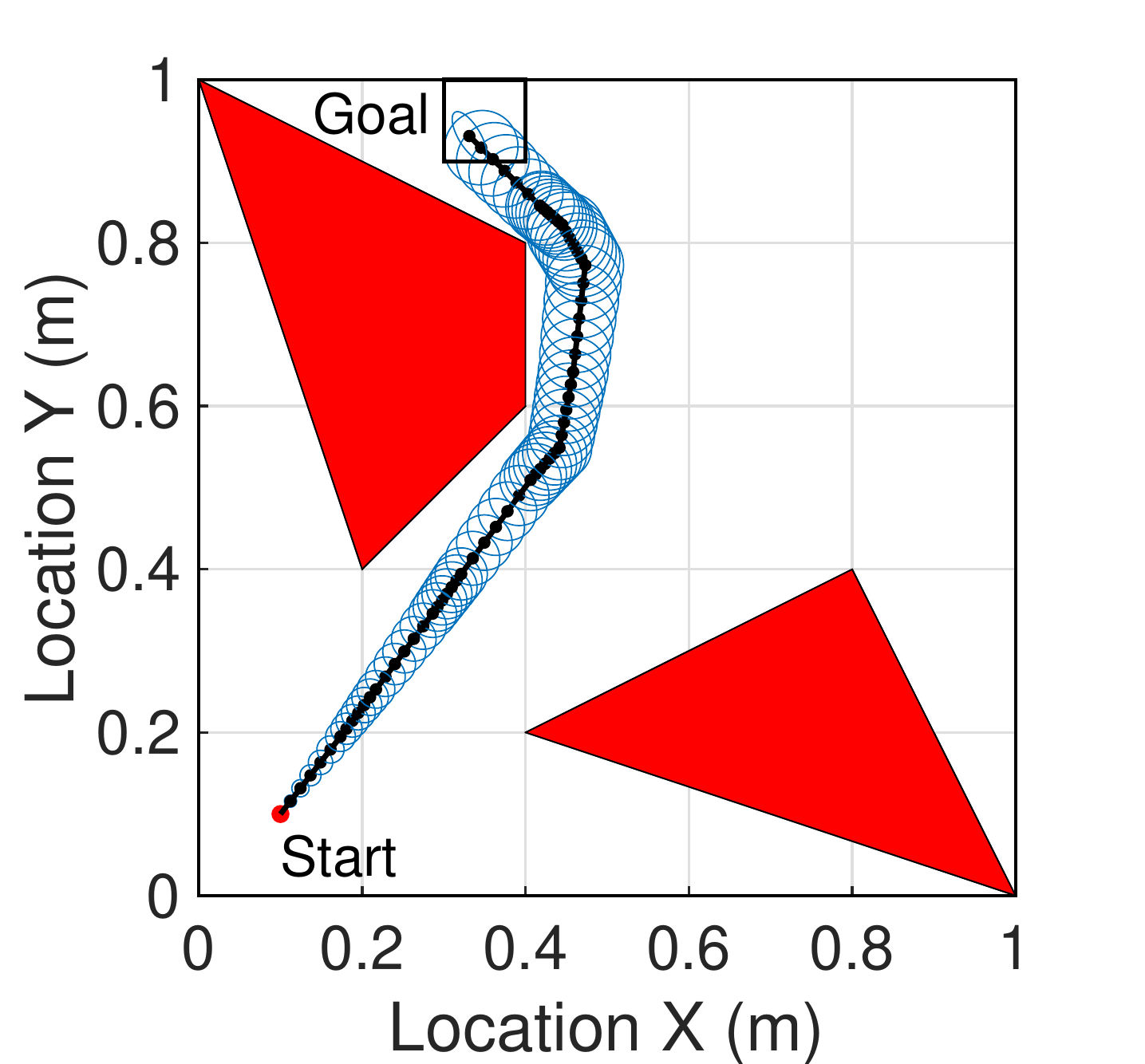} &\!\!\!\!\!\!\!\!\!\!\!\!\!\!\! \includegraphics[scale=0.3]{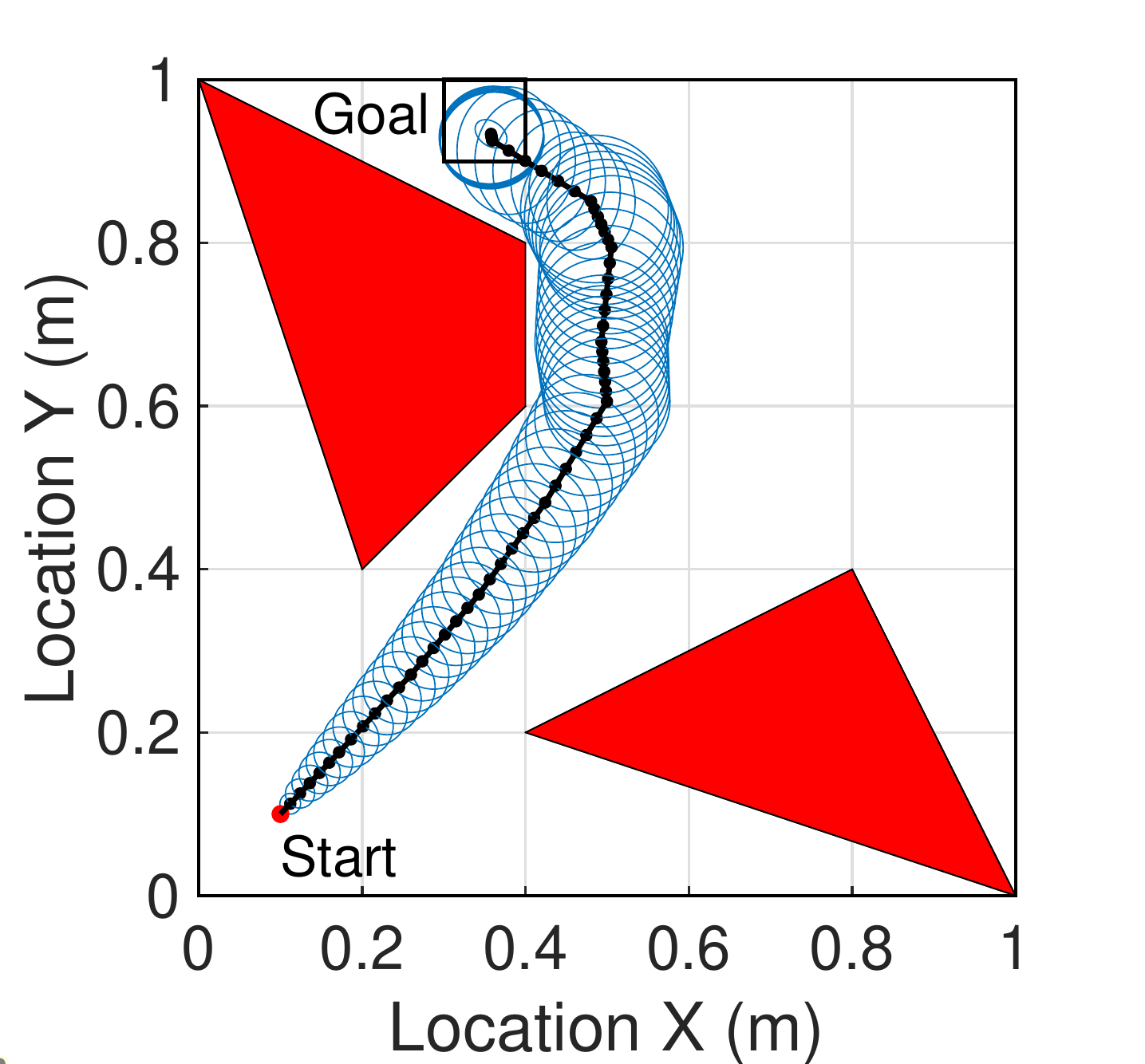}\\
      (c)&(d)
       \end{tabular}
        \caption{Trajectories planned with the instantaneous safety criterion \cite{pedram2021rationally}. The red-faced polygons represent $\chi_{obs}$. The trajectories are shown with (a) $25\%$, (b) $50\%$, (c) $75\%$ and (d) $99\%$ confidence ellipses at all the time steps.}
        \label{Fig. planned trajectories}
\end{figure}
\begin{figure}[tbhp]
    \centering
    \includegraphics[scale=0.23]{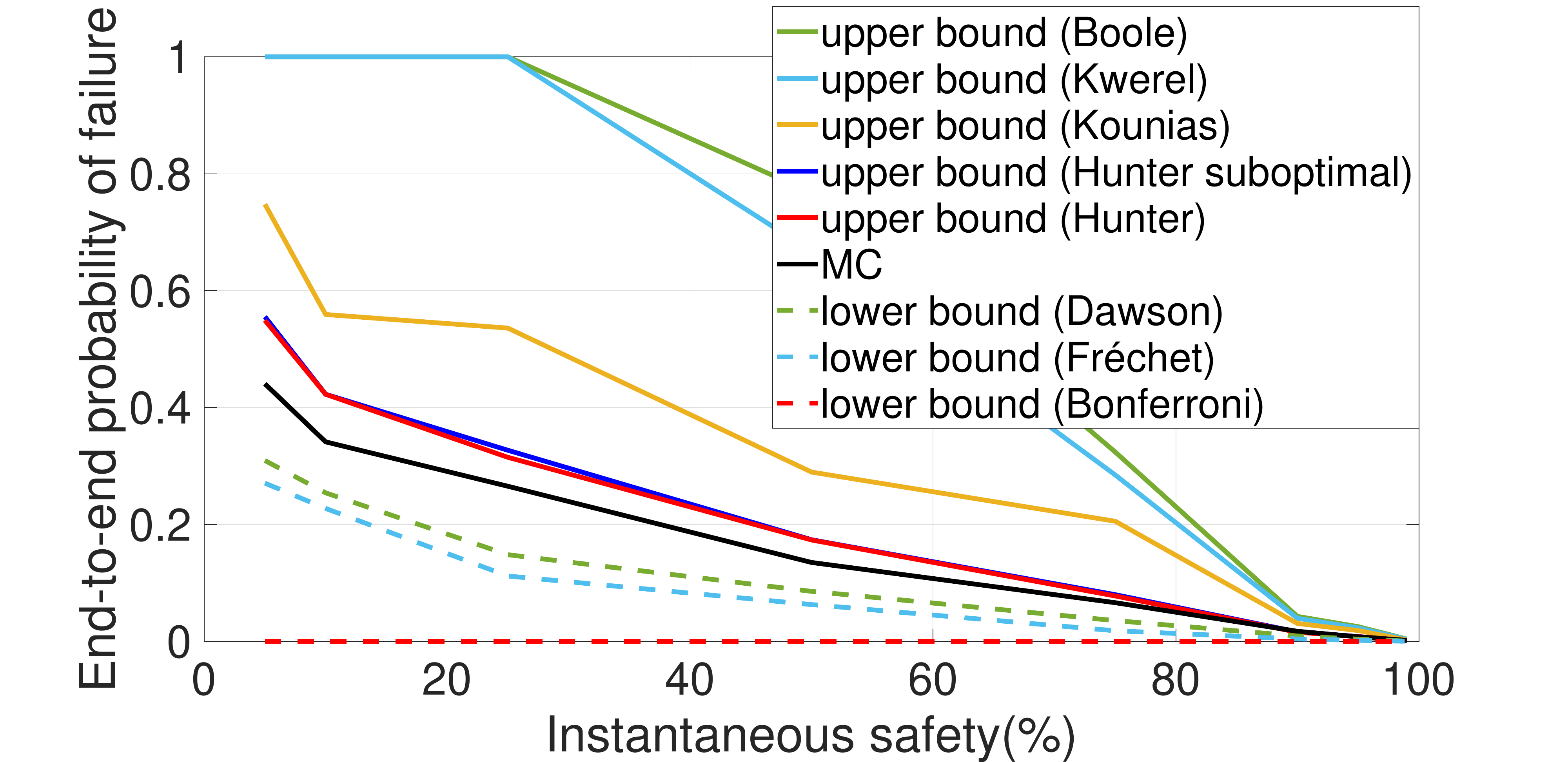}
    \caption{End-to-end probabilities of failure estimated by Monte Carlo simulations (plotted in black) and their analytical bounds for the trajectories generated with different instantaneous safety levels.}
    \label{Fig. all bounds}
\end{figure}

First, we plan the trajectories using RRT* with instantaneous safety criterion \cite{pedram2021rationally} (i.e., at every time step, the confidence ellipse with a fixed safety level is collision-free) and compute our bounds for these plans. For a given configuration space, four planned trajectories with $25\%$, $50\%$, $75\%$, and $99\%$ instantaneous safety levels are shown in Fig. \ref{Fig. planned trajectories} respectively and our bounds for the end-to-end probabilities of failure vs instantaneous safety levels are plotted in Fig. \ref{Fig. all bounds}. 
We validate our bounds by comparing them with the failure probabilities computed using $10^5$ Monte Carlo simulations (shown in black). Each trajectory execution of a Monte Carlo simulation is sampled from the distribution of $\overline{x}_{traj}$ given in (\ref{X_traj}). Bonferroni's second-order lower bounds (shown with red dashed line) are trivial for all the paths in this example. As evident from the graph, Hunter's upper bound or its suboptimal version and Dawson and Sankoff's lower bound together provide close approximation to the Monte Carlo estimates of the end-to-end probability of failure. The graph shows that the bounds presented in this work are significantly less conservative than Boole's bound.
\par
Next, we demonstrate a larger statistical evaluation over $100$ trajectories planned using RRT* in randomly-generated environments (random initial, goal and obstacle positions). These trajectories are nominally safe i.e., only the planned positions $\{x_t^{plan}\}_{t=0, 1, \hdots, T}$ are ensured to be collision-free. Table \ref{Table: comparison} compares the mean absolute errors of different bounds with respect to $10^5$ Monte Carlo simulations. 
\begin{table}[h]
\caption{Comparison of different bounds over $100$ trajectories in terms of Mean Absolute Error with respect to $10^5$ Monte Carlo simulations and computation time. Computation is performed in MATLAB on a consumer laptop.}
\label{Table: comparison}
\begin{center}
\begin{tabular}{ |c|c|c| } 
 \hline
 \textbf{Estimates} & \textbf{Mean Absolute Error [\%]} & \textbf{Avg. Time [s]} \\ 
 \hline\hline
  Monte Carlo & 0 & 46.83\\  
  \hline
  \textit{Upper bounds} &&\\
 Boole & 40.59 & 0.01   \\
 Kwerel & 38.15 &  2.43  \\
 Kounias & 13.34 & 2.42   \\
 Hunter & 8.63 &  2.41  \\
 Hunter suboptimal & 10.25 & 0.18   \\
 \hline
 \textit{Lower bounds} &&\\
 Bonferroni & 54.88  & 2.44   \\
 Fr\'echet & 40.08 &  0.01  \\
 Dawson & 16.74 & 2.44  \\
 \hline
\end{tabular}
\end{center}
\end{table}
If the purpose of risk estimation is verification and performance analysis, then it can be performed off-line. However, when risk estimation is a part of online motion planning algorithms, its computation time plays an important role. The computation times for our MATLAB implementation of these bounds and the Monte Carlo method are also reported in Table \ref{Table: comparison}. From the data presented, we can draw the following conclusions. First, the bounds presented in this work require significantly less computation time as compared to the Monte Carlo method. Second, our upper bounds provide considerably tighter estimates than Boole's bound at the expense of some additional computational overhead. Dawson and Hunter's estimates provide respectively the best lower and upper bounds of risk among all. Finally, Hunter's suboptimal bound even though slightly more conservative, is computationally cheaper than Hunter's bound. As it also possesses the time-additive structure, this bound could be embedded in the risk-aware motion planning framework.

\section{DISCUSSION} \label{Sec: DISCUSSION} 
\subsection{Risk Bounds for Continuous-Time Systems}\label{subsection: Continuous-Time End-to-End Risks}
\begin{figure}[tbhp]
      \centering
      \includegraphics[scale=0.23]{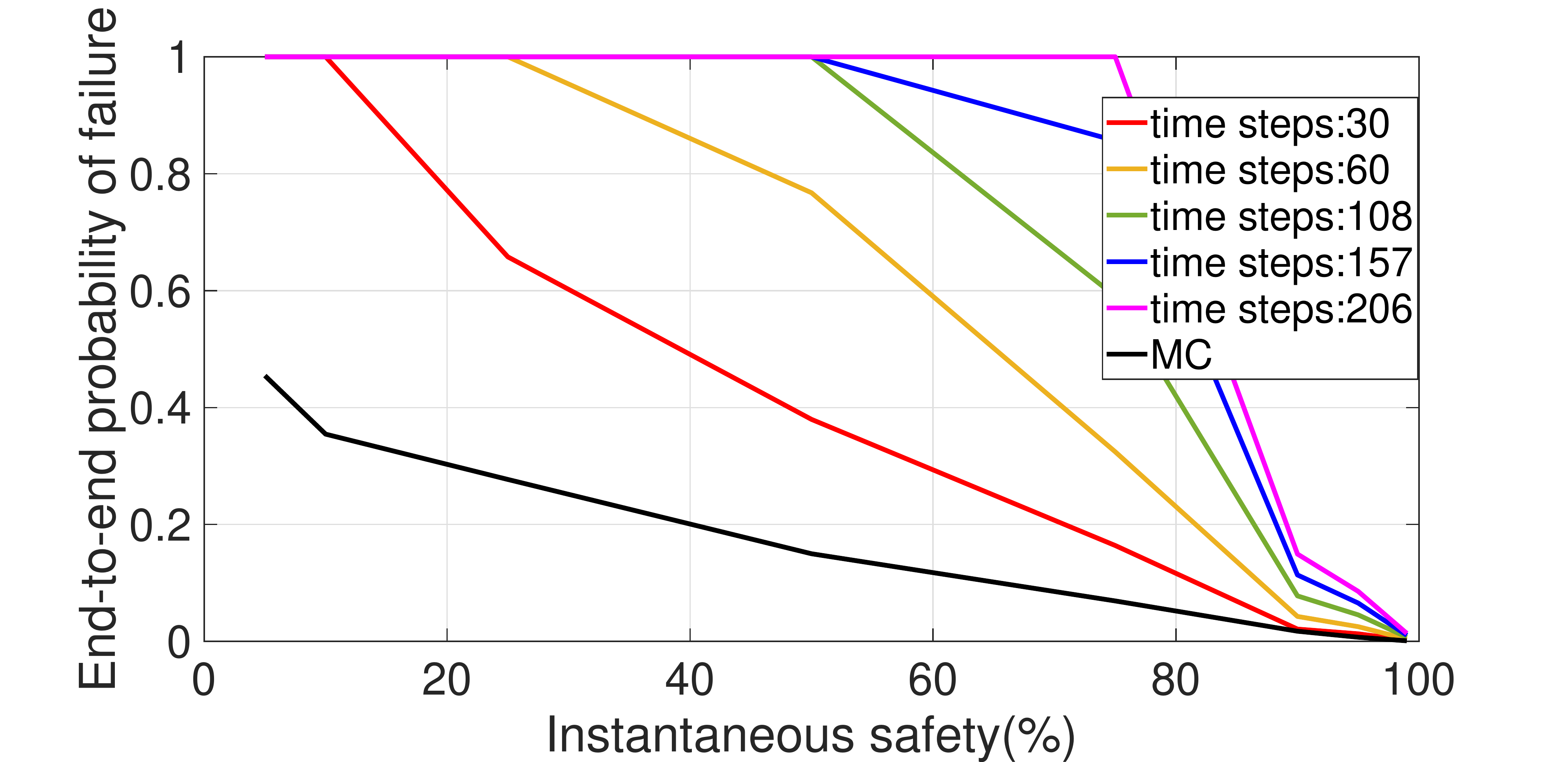}
      \caption{Boole's upper bounds of end-to-end probabilities of failure for different sampling rates vs instantaneous safety levels. The black graph shows the end-to-end probabilities of failure estimated by Monte Carlo simulations for time steps: $206$.}
      \label{Fig. Boole convergence}
   \end{figure}

\begin{figure}[tbhp]
      \centering
      \includegraphics[scale=0.23]{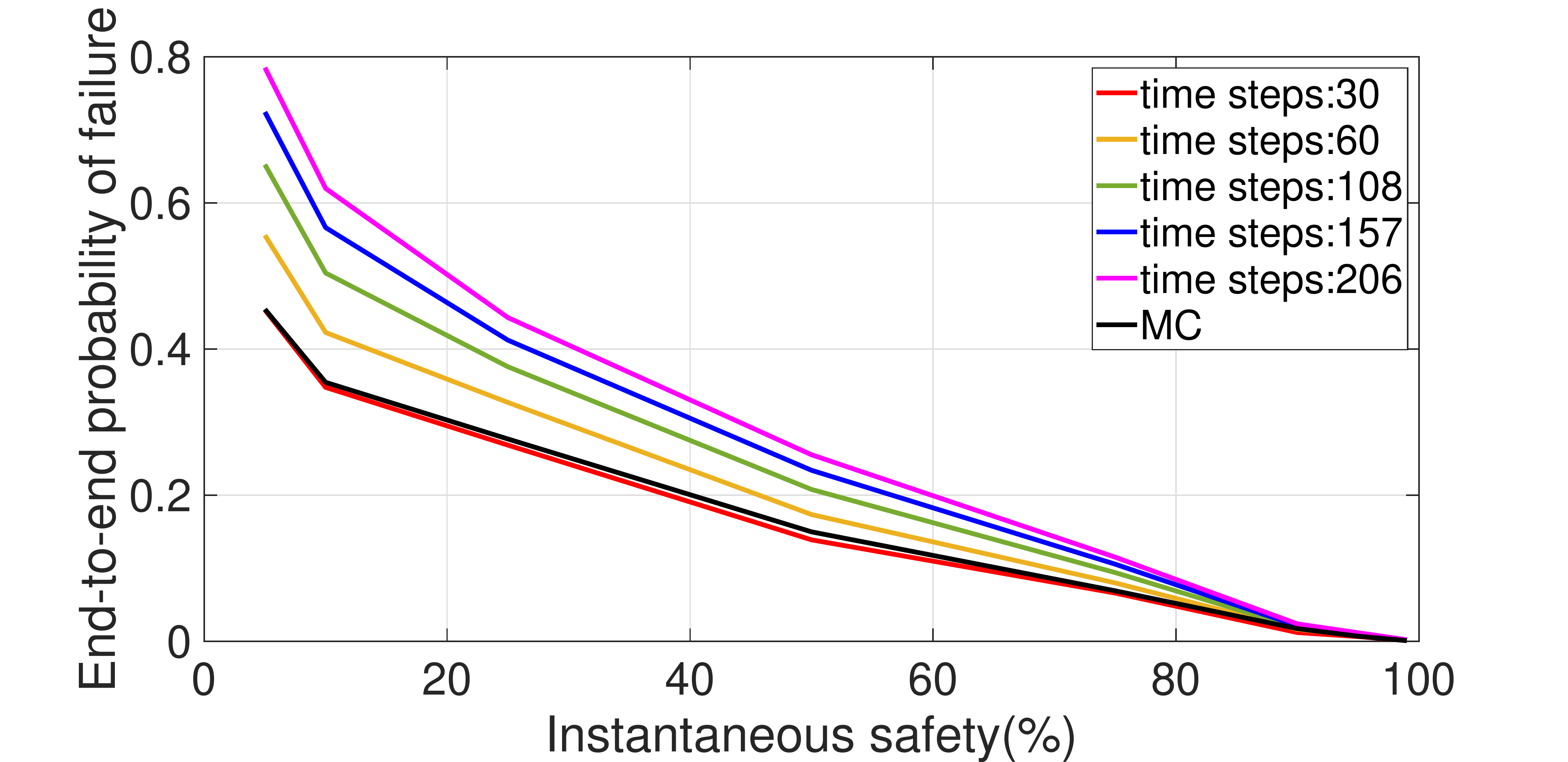}
      \caption{Hunter's suboptimal upper bounds of end-to-end probabilities of failure for  different sampling rates vs instantaneous safety levels. The black graph shows the end-to-end probabilities of failure estimated by Monte Carlo simulations for time steps: $206$.}
      \label{Fig. Hunter convergence}
   \end{figure}

\begin{figure}[tbhp]
      \centering
      \includegraphics[scale=0.6]{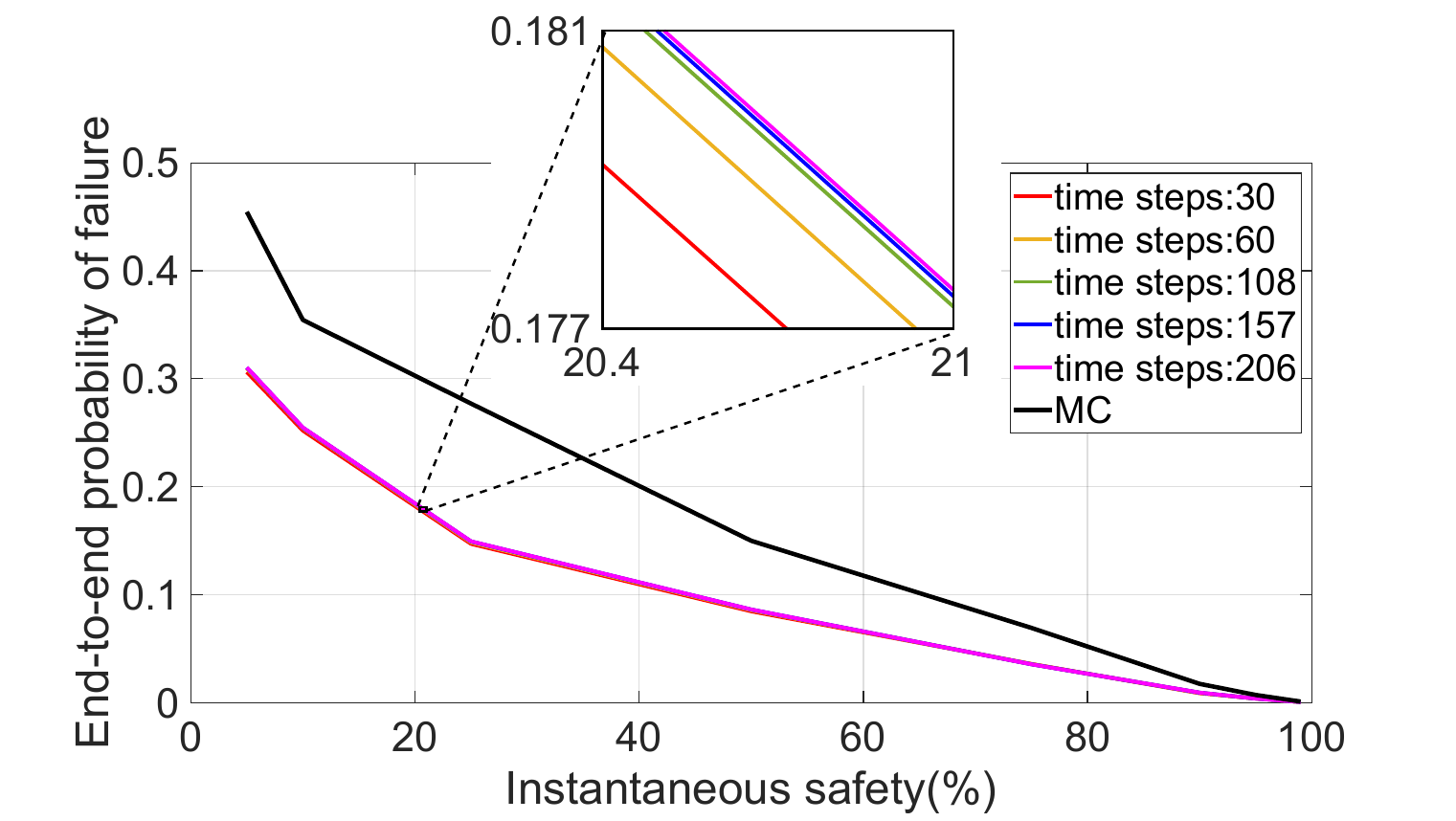}
      \caption{Dawson and Sankoff's lower bounds of end-to-end probabilities of failure for different sampling rates vs instantaneous safety levels. The black graph shows the end-to-end probabilities of failure estimated by Monte Carlo simulations for time steps: $206$. The zoomed view of a small portion of the graph is shown.}
      \label{Fig. Dawson convergence}
   \end{figure}

Although our results so far are restricted to discrete-time systems, in practice we are often interested in the safety of continuous-time systems. Hence, it is of our natural interest to study the impact of increased sampling rates on the aforementioned bounds and how they can be used to imply the safety of continuous-time systems. Consider the configuration space and trajectories from the first example of Section \ref{Sec: EXAMPLE}. Assume the trajectories are planned for the continuous-time system and its time discretization yields (\ref{eg true dynamics}). For this system, the probability bounds of Boole, suboptimal Hunter, and Dawson at different rates of time discretization vs instantaneous safety levels are plotted in Fig. \ref{Fig. Boole convergence}, \ref{Fig. Hunter convergence} and \ref{Fig. Dawson convergence} respectively. The probabilities obtained using Monte Carlo simulations for a high rate of time discretization (time steps: $206$) are plotted in black, in all three figures. The lower bounds for the discrete-time risks at all the rates of time-discretization should also be valid for the continuous-time risks. Moreover, Fig. \ref{Fig. Dawson convergence} shows that Dawson and Sankoff's lower bound becomes sharper with the increase in the sampling rate. Similarly, it can be shown that Fr\'echet's bound also increases in sharpness at the higher sampling rates. On the contrary, Fig. \ref{Fig. Boole convergence} and \ref{Fig. Hunter convergence} show that Boole's and Hunter's suboptimal upper bounds decrease in sharpness with the increase in the sampling rate. However, there is a remarkable difference in the rates at which they lose sharpness. Boole's bound quickly diverges to the trivial probability of $1$ as the sampling rate is increased, unlike the suboptimal Hunter's bound. It can be shown that Hunter's and Kounias' bounds also lose sharpness at the higher sampling rates but they still perform better than Boole's bound. Hence, these inequalities could be useful even for the systems operated in continuous-time.
\par 
More investigation and comparison of our bounds at the high sampling rates with the continuous-time risk estimates computed in the existing literature \cite{shah2011probability}-\cite{oguri2019convex} are left for the future work. Another direction of using these bounds for the continuous-time settings could be similar to \cite{ariu2017chance}, in which the authors use the reflection principle and apply Boole's inequality over intervals instead of the discrete-time steps to compute risk bounds in continuous-time. The reflection principle similar to \cite{ariu2017chance} could be used with the new bounds presented in this work to obtain less conservative risk estimates for the continuous-time models. 
   \par
\subsection{Higher-Order Probability Bounds}
In this work, we have implemented first and second-order probability bounds. The question naturally arises whether we can consider bounds of order higher than two. We have seen Bonferroni's first and second-order bounds in (\ref{Bonferroni's first order upper bound}) and (\ref{Bonferroni's second order lower bound}), respectively. The classical inclusion-exclusion principle states that 
\begin{equation}\label{principle of inclusion and exclusion}
   P\left(E\right)=S_1-S_2+S_3-S_4+\hdots+\left(-1\right)^TS_{T+1},  
\end{equation}
where $S_1$ and $S_2$ are given by (\ref{S_1,S_2}). In general, $S_r$, $1\leq r\leq T+1$, is defined as
\begin{equation*}\label{S_r}
    S_r\coloneqq\sum\limits_{0\leq j_1<\hdots<j_r\leq T}\!\!\!\!\!\!\!P\left(E_{j_1}\bigcap\hdots \bigcap E_{j_r}\right).
\end{equation*}
The sum of the first $r$ terms on the right side of (\ref{principle of inclusion and exclusion}) provides an upper bound to $P\left(E\right)$ when $r$ is odd and a lower bound when $r$ is even, producing Bonferroni's $r^{th}$-order bound. However, it is generally not true that Bonferroni's bounds increase in sharpness with the order \cite{schwager1984bonferroni}. Hence, Bonferroni's higher-order inequalities might not give sharper bounds than the ones considered in this work. A third-order upper bound computed using the Cherry Trees approach presented in \cite{bukszar2001probability} can be sharper than Hunter's upper bound. Sharper higher-order upper and lower bounds can be computed using the linear programming algorithms \cite{prekopa1988boole}, \cite{prekopa2003probabilistic}. Of course, higher-order bounds are associated with higher computational complexities.  
\section{CONCLUSION} \label{Sec: CONCLUSION}
In this work, we presented several upper and lower bounds for the probability of collision while tracking a given motion plan under stochastic uncertainties. Our approach makes no independence assumptions on the events of collision at different time steps and computes less conservative bounds for the failure probability than the commonly used Boole’s bound in the literature. The approach is quite general and can be applied to any discrete-time trajectory tracking scenario regardless of the choice of linear feedback control laws and trajectory generation algorithms. We also study the performance of the derived discrete-time risk bounds in the continuous-time setting and show that our bounds perform better than Boole's bound even when the underlying discretization is refined. The future work includes the incorporation of these bounds in planning algorithms to generate risk-optimal trajectories. 



\section*{APPENDIX} 
\section*{Computation of $p_t$ and $p_{s, t}$ when obstacles are polyhedral}
\begin{figure}[tbhp]
      \centering
      \includegraphics[scale=0.65]{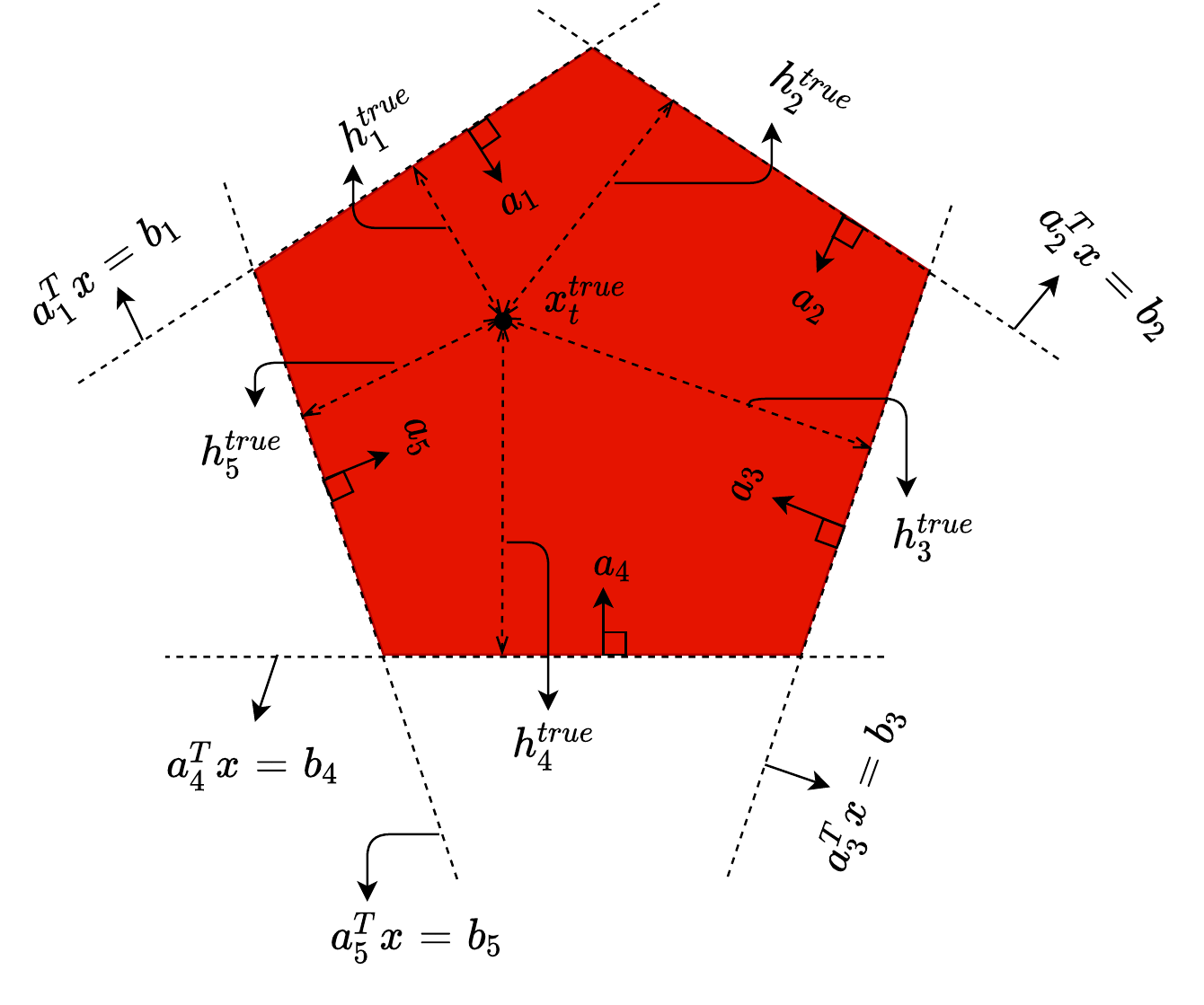}
      \caption{The polyhedral obstacle, $\chi_{obs_l}$, composed of $5$ linear constraints. The black dot represents $x_t^{true}$. The perpendicular distances between $x_t^{true}$ and the linear constraints are also shown. For convenience, the subscript $l$ is removed from $a_{i,l}$, $b_{i,l}$ and $h^{true}_{i,l}$. }
      \label{Fig. obstacle_figure}
   \end{figure}

Assume that the obstacle region $\chi_{obs}$ is decomposed into the disjoint union of $L$ polyhedrons, $\chi_{obs_l}$, $1\leq l\leq L$. $\chi_{obs_l}$ can be represented as a conjunction of $I_l$ linear constraints as follows:
\begin{equation}\label{chi_obs_l}
    \chi_{obs_l}=\bigcap\limits_{i=1}^{I_l}\left(x:a^T_{i,l}x\geq b_{i,l}\right),\qquad \forall\;1\leq l\leq L.
\end{equation}
The vector $a_{i,l}$ is the unit normal of the constraint $i$ of the polyhedron $l$, pointing inside the polyhedron. Define
\begin{equation}\label{a_obs_l}
 a_{obs_l}\coloneqq \begin{bmatrix} a_{1,l}&\hdots&a_{I_l,l}
 \end{bmatrix}, \qquad \forall\;1\leq l\leq L.
\end{equation}
Let $h_{i,l}^{true}$ be a univariate random variable corresponding to the perpendicular distance between the constraint $i$ of the polyhedron $l$ and $x_t^{true}$ as shown in Fig. \ref{Fig. obstacle_figure}. It can be shown that \cite{blackmore2011chance}
\begin{equation*}
   h_{i,l}^{true}\sim\mathcal{N}\left(h_{i,l}^{plan},H_{i,l}\right),\qquad \forall\;1\leq l\leq L,
\end{equation*}
where 
\begin{equation*}
 h_{i,l}^{plan}\coloneqq a_{i,l}^Tx_t^{plan}-b_{i,l}\quad\text{and}\quad H_{i,l}\coloneqq a_{i,l}^TX_ta_{i,l}.  
\end{equation*}
\subsection{Computation of $p_t$}
We can write $p_t$ as
\begin{equation}\label{p_t as sum of p_tl}
    p_t=\sum\limits_{l=1}^{L}p_{t,l},
\end{equation}
where
\begin{equation}\label{P_tl with chi_obs}
    p_{t,l}=P\left(x_t^{true}\in\chi_{obs_l}\right).
\end{equation}
Using (\ref{chi_obs_l}), (\ref{P_tl with chi_obs}) can be written as 
\begin{equation}\label{P_tl with a and b}
    p_{t,l}=P\left(\bigcap\limits_{i=1}^{I_l}a_{i,l}^Tx_t^{true}\geq b_{i,l}\right).
\end{equation}
The event $a_{i,l}^Tx_t^{true}\geq b_{i,l}$ is equivalent to $h_{i,l}^{true}\geq0$. Hence, (\ref{P_tl with a and b}) can be written as 
\begin{equation*}
   p_{t,l}=P\left(\bigcap\limits_{i=1}^{I_l}h_{i,l}^{true}\geq0\right). 
\end{equation*}
Defining 
\begin{equation*}
 h_{obs_l}^{true}\coloneqq\begin{bmatrix}
 h_{1,l}^{true}&\hdots&h_{I_l,l}^{true}
 \end{bmatrix}^T,
\end{equation*}
it can be shown that 
\begin{equation*}
    h_{obs_l}^{true}\sim\mathcal{N}\left(h_{obs_l}^{plan},H_{obs_l}\right),
\end{equation*}
where
\begin{equation*}
    h_{obs_l}^{plan}\coloneqq\begin{bmatrix}
 h_{1,l}^{plan}&\hdots&h_{I_l,l}^{plan}
 \end{bmatrix}^T
\end{equation*}
and
\begin{equation*}
    H_{obs_l}\coloneqq a_{obs_l}^TX_ta_{obs_l},
\end{equation*}
where $a_{obs_l}$ is defined as (\ref{a_obs_l}).
Then, $p_{t,l}$ can be computed as 
\begin{equation}\label{p_tl final}
    p_{t,l}=\int_{0}^{\infty}\hdots \int_{0}^{\infty}\mathcal{N}\left(h_{obs_l}^{plan},H_{obs_l}\right)h_{1,l}^{true}\hdots h_{I_l,l}^{true}.
\end{equation}
Computing the integral (\ref{p_tl final}) is much easier than evaluating the integral (\ref{p_t}). We use MATLAB’s \verb_mvncdf_ function to compute the integral (\ref{p_tl final}) over the hypercube. 
\subsection{Computation of $p_{s, t}$}
Similarly to (\ref{p_t as sum of p_tl}), we can write $p_{s, t}$ as
\begin{equation*}
    p_{s, t}=\sum\limits_{l=1}^{L}\sum\limits_{m=1}^{L}p_{st,lm},
\end{equation*}
where
\begin{equation}\label{P_st,lm with chi_obs}
  p_{st,lm}=P\left(\left(x_s^{true}\in\chi_{obs_l}\right)\bigcap\left(x_t^{true}\in\chi_{obs_m}\right)\right).  
\end{equation}
Using (\ref{chi_obs_l}), (\ref{P_st,lm with chi_obs}) can be written as 
\begin{equation}\label{P_stlm with a and b}
\begin{split}
    p_{st,lm}=&\\
    &\!\!\!\!\!\!\!\!\!\!\!\!\!\!\!\!\!\!\!\!\!\!\!P\left(\left(\bigcap\limits_{i=1}^{I_l}a_{i,l}^Tx_s^{true}\geq b_{i,l}\right)\bigcap\left(\bigcap\limits_{i=1}^{I_m}a_{i,m}^Tx_t^{true}\geq b_{i,m}\right)\right).
\end{split}
\end{equation}
The events $a_{i,l}^Tx_s^{true}\geq b_{i,l}$ and $a_{i,m}^Tx_t^{true}\geq b_{i,m}$ are equivalent to $h_{i,l}^{true}\geq0$ and $h_{i,m}^{true}\geq0$ respectively. Hence, (\ref{P_stlm with a and b}) can be written as 
\begin{equation*}
   p_{st,lm}=P\left(\left(\bigcap\limits_{i=1}^{I_l}h_{i,l}^{true}\geq0\right)\bigcap\left(\bigcap\limits_{i=1}^{I_m}h_{i,m}^{true}\geq0\right)\right). 
\end{equation*}
Defining 
\begin{equation*}
 h_{obs_{lm}}^{true}\coloneqq\begin{bmatrix}
 h_{1,l}^{true}&\hdots&h_{I_l,l}^{true}&h_{1,m}^{true}&\hdots&h_{I_m,m}^{true}
 \end{bmatrix}^T,
\end{equation*}
it can be shown that 
\begin{equation*}
    h_{obs_{lm}}^{true}\sim\mathcal{N}\left(h_{obs_{lm}}^{plan},H_{obs_{lm}}\right),
\end{equation*}
where
\begin{equation*}
   h_{obs_{lm}}^{plan}\coloneqq\begin{bmatrix}
 h_{1,l}^{plan}&\hdots&h_{I_l,l}^{plan}&h_{1,m}^{plan}&\hdots&h_{I_m,m}^{plan}
 \end{bmatrix}^T
\end{equation*}
and
\begin{equation*}
    H_{obs_{lm}}\coloneqq\begin{bmatrix}
    a_{obs_l}^TK_{ss}a_{obs_l}&a_{obs_l}^TK_{st}a_{obs_m}\\
    a_{obs_m}^TK_{st}^Ta_{obs_l}&a_{obs_m}^TK_{tt}a_{obs_m}
 \end{bmatrix},
\end{equation*}
where $a_{obs_l}$ and $a_{obs_m}$ are defined as (\ref{a_obs_l}) and
\begin{equation}
    K_{ss}\coloneqq X_s,\;K_{tt}\coloneqq X_t,\; K_{st}\coloneqq cov\left(x_s^{true},x_t^{true}\right).
\end{equation}

Then, $p_{st,lm}$ can be computed as 
\begin{equation}\label{p_stlm final}
\begin{split}
 p_{st,lm}=&\\
 &\!\!\!\!\!\!\!\!\!\!\!\!\!\!\!\!\!\!\!\!\!\!\!\!\int_{0}^{\infty}\!\!\!\!\!\!\hdots\! \int_{0}^{\infty}\!\!\!\!\!\mathcal{N}\!\left(\!h_{obs_{lm}}^{plan},H_{obs_{lm}}\!\right)\!h_{1,l}^{true}\!\!\hdots h_{I_l,l}^{true}h_{1,m}^{true}\!\hdots h_{I_m,m}^{true}.   
\end{split}
\end{equation}
Again, the MATLAB’s \verb_mvncdf_ function can be used to compute the integral (\ref{p_stlm final}) over the hypercube. 
\section*{ACKNOWLEDGMENT}
We would like to thank Ali Reza Pedram for the helpful suggestions on the probability bounds. 
\bibliographystyle{IEEEtran}
\bibliography{bibliography}

\begin{thebibliography}{10}
\providecommand{\url}[1]{#1}
\csname url@rmstyle\endcsname
\providecommand{\newblock}{\relax}
\providecommand{\bibinfo}[2]{#2}
\providecommand\BIBentrySTDinterwordspacing{\spaceskip=0pt\relax}
\providecommand\BIBentryALTinterwordstretchfactor{4}
\providecommand\BIBentryALTinterwordspacing{\spaceskip=\fontdimen2\font plus
\BIBentryALTinterwordstretchfactor\fontdimen3\font minus
  \fontdimen4\font\relax}
\providecommand\BIBforeignlanguage[2]{{%
\expandafter\ifx\csname l@#1\endcsname\relax
\typeout{** WARNING: IEEEtran.bst: No hyphenation pattern has been}%
\typeout{** loaded for the language `#1'. Using the pattern for}%
\typeout{** the default language instead.}%
\else
\language=\csname l@#1\endcsname
\fi
#2}}

\bibitem{lavalle2006planning}
S.~M. LaValle, \emph{Planning algorithms}.\hskip 1em plus 0.5em minus
  0.4em\relax Cambridge university press, 2006.

\bibitem{pepy2006safe}
R.~Pepy and A.~Lambert, ``Safe path planning in an uncertain-configuration
  space using {RRT},'' in \emph{2006 IEEE/RSJ International Conference on
  Intelligent Robots and Systems}.\hskip 1em plus 0.5em minus 0.4em\relax IEEE,
  2006, pp. 5376--5381.

\bibitem{blackmore2011chance}
L.~Blackmore, M.~Ono, and B.~C. Williams, ``Chance-constrained optimal path
  planning with obstacles,'' \emph{IEEE Transactions on Robotics}, vol.~27,
  no.~6, pp. 1080--1094, 2011.

\bibitem{strawser2018approximate}
D.~Strawser and B.~Williams, ``Approximate branch and bound for fast,
  risk-bound stochastic path planning,'' in \emph{2018 IEEE International
  Conference on Robotics and Automation (ICRA)}.\hskip 1em plus 0.5em minus
  0.4em\relax IEEE, 2018, pp. 7047--7054.

\bibitem{majumdar2013robust}
A.~Majumdar and R.~Tedrake, ``Robust online motion planning with regions of
  finite time invariance,'' in \emph{Algorithmic foundations of robotics
  X}.\hskip 1em plus 0.5em minus 0.4em\relax Springer, 2013, pp. 543--558.

\bibitem{lopez2019dynamic}
B.~T. Lopez, J.-J.~E. Slotine, and J.~P. How, ``Dynamic tube {MPC} for
  nonlinear systems,'' in \emph{2019 American Control Conference (ACC)}.\hskip
  1em plus 0.5em minus 0.4em\relax IEEE, 2019, pp. 1655--1662.

\bibitem{stengel1994optimal}
R.~F. Stengel, \emph{Optimal control and estimation}.\hskip 1em plus 0.5em
  minus 0.4em\relax Courier Corporation, 1994.

\bibitem{schouwenaars2001mixed}
T.~Schouwenaars, B.~De~Moor, E.~Feron, and J.~How, ``Mixed integer programming
  for multi-vehicle path planning,'' in \emph{2001 European control conference
  (ECC)}.\hskip 1em plus 0.5em minus 0.4em\relax IEEE, 2001, pp. 2603--2608.

\bibitem{schouwenaars2001plume}
T.~Schouwenaars, A.~Richards, E.~Feron, and J.~How, ``Plume avoidance maneuver
  planning using mixed integer linear programming,'' in \emph{AIAA Guidance,
  Navigation, and Control Conference and Exhibit}, 2001, p. 4091.

\bibitem{blackmore2010probabilistic}
L.~Blackmore, M.~Ono, A.~Bektassov, and B.~C. Williams, ``A probabilistic
  particle-control approximation of chance-constrained stochastic predictive
  control,'' \emph{IEEE transactions on Robotics}, vol.~26, no.~3, pp.
  502--517, 2010.

\bibitem{janson2018monte}
L.~Janson, E.~Schmerling, and M.~Pavone, ``Monte {C}arlo motion planning for
  robot trajectory optimization under uncertainty,'' in \emph{Robotics
  Research}.\hskip 1em plus 0.5em minus 0.4em\relax Springer, 2018, pp.
  343--361.

\bibitem{calafiore2006scenario}
G.~C. Calafiore and M.~C. Campi, ``The scenario approach to robust control
  design,'' \emph{IEEE Transactions on automatic control}, vol.~51, no.~5, pp.
  742--753, 2006.

\bibitem{ono2008efficient}
M.~Ono and B.~C. Williams, ``An efficient motion planning algorithm for
  stochastic dynamic systems with constraints on probability of failure.'' in
  \emph{AAAI}, 2008, pp. 1376--1382.

\bibitem{ono2015chance}
M.~Ono, M.~Pavone, Y.~Kuwata, and J.~Balaram, ``Chance-constrained dynamic
  programming with application to risk-aware robotic space exploration,''
  \emph{Autonomous Robots}, vol.~39, no.~4, pp. 555--571, 2015.

\bibitem{blackmore2009convex}
L.~Blackmore and M.~Ono, ``Convex chance constrained predictive control without
  sampling,'' in \emph{AIAA Guidance, Navigation, and Control Conference},
  2009, p. 5876.

\bibitem{patil2012estimating}
S.~Patil, J.~Van Den~Berg, and R.~Alterovitz, ``Estimating probability of
  collision for safe motion planning under {G}aussian motion and sensing
  uncertainty,'' in \emph{2012 IEEE International Conference on Robotics and
  Automation}.\hskip 1em plus 0.5em minus 0.4em\relax IEEE, 2012, pp.
  3238--3244.

\bibitem{frey2020collision}
K.~M. Frey, T.~J. Steiner, and J.~How, ``Collision probabilities for
  continuous-time systems without sampling,'' \emph{Proceedings of Robotics:
  Science and Systems. Corvalis, Oregon, USA (July 2020)}, 2020.

\bibitem{kwerel1975most}
S.~M. Kwerel, ``Most stringent bounds on aggregated probabilities of partially
  specified dependent probability systems,'' \emph{Journal of the American
  Statistical Association}, vol.~70, no. 350, pp. 472--479, 1975.

\bibitem{kwerel1975bounds}
------, ``Bounds on the probability of the union and intersection of m
  events,'' \emph{Advances in Applied Probability}, vol.~7, no.~2, pp.
  431--448, 1975.

\bibitem{sathe1980inequalities}
Y.~Sathe, M.~Pradhan, and S.~Shah, ``Inequalities for the probability of the
  occurrence of at least m out of n events,'' \emph{Journal of Applied
  Probability}, vol.~17, no.~4, pp. 1127--1132, 1980.

\bibitem{kruskal1956shortest}
J.~B. Kruskal, ``On the shortest spanning subtree of a graph and the traveling
  salesman problem,'' \emph{Proceedings of the American Mathematical society},
  vol.~7, no.~1, pp. 48--50, 1956.

\bibitem{prekopa1988boole}
A.~Pr{\'e}kopa, ``Boole-{B}onferroni inequalities and linear programming,''
  \emph{Operations Research}, vol.~36, no.~1, pp. 145--162, 1988.

\bibitem{dawson1967inequality}
D.~Dawson and D.~Sankoff, ``An inequality for probabilities,''
  \emph{Proceedings of the American Mathematical Society}, vol.~18, no.~3, pp.
  504--507, 1967.

\bibitem{galambos1977bonferroni}
J.~Galambos, ``Bonferroni inequalities,'' \emph{The Annals of Probability}, pp.
  577--581, 1977.

\bibitem{pedram2021rationally}
A.~R. Pedram, J.~Stefarr, R.~Funada, and T.~Tanaka, ``Rationally inattentive
  path-planning via {RRT},'' in \emph{2021 American Control Conference
  (ACC)}.\hskip 1em plus 0.5em minus 0.4em\relax IEEE, 2021, pp. 3440--3446.

\bibitem{shah2011probability}
S.~K. Shah, C.~D. Pahlajani, and H.~G. Tanner, ``Probability of success in
  stochastic robot navigation with state feedback,'' in \emph{2011 IEEE/RSJ
  International Conference on Intelligent Robots and Systems}.\hskip 1em plus
  0.5em minus 0.4em\relax IEEE, 2011, pp. 3911--3916.

\bibitem{oguri2019convex}
K.~Oguri, M.~Ono, and J.~W. McMahon, ``Convex optimization over sequential
  linear feedback policies with continuous-time chance constraints,'' in
  \emph{2019 IEEE 58th Conference on Decision and Control (CDC)}.\hskip 1em
  plus 0.5em minus 0.4em\relax IEEE, 2019, pp. 6325--6331.

\bibitem{ariu2017chance}
K.~Ariu, C.~Fang, M.~Arantes, C.~Toledo, and B.~Williams, ``Chance-constrained
  path planning with continuous time safety guarantees,'' in \emph{Workshops at
  the Thirty-First AAAI Conference on Artificial Intelligence}, 2017.

\bibitem{schwager1984bonferroni}
S.~J. Schwager, ``Bonferroni sometimes loses,'' \emph{The American
  Statistician}, vol.~38, no.~3, pp. 192--197, 1984.

\bibitem{bukszar2001probability}
J.~Buksz{\'a}r and A.~Prekopa, ``Probability bounds with cherry trees,''
  \emph{Mathematics of Operations Research}, vol.~26, no.~1, pp. 174--192,
  2001.

\bibitem{prekopa2003probabilistic}
A.~Pr\'ekopa, ``Probabilistic programming,'' \emph{Handbooks in operations
  research and management science}, vol.~10, pp. 267--351, 2003.

\end{thebibliography}
\end{document}